\documentclass[letterpaper, 10 pt, conference]{ieeeconf}  

\IEEEoverridecommandlockouts                              

\usepackage{graphics} 
\usepackage{graphicx} 
\usepackage{amsmath} 
\usepackage[noend]{algpseudocode}
\usepackage{algorithmicx,algorithm}
\usepackage{cite}
\usepackage{multirow}
\usepackage{hyperref}
\usepackage{mathrsfs}
\usepackage{bbm}
\usepackage{color}
\usepackage{tabularx}
\usepackage{threeparttable}
\usepackage{amssymb}
\usepackage{xspace}
\usepackage{float}
\usepackage{booktabs}
\usepackage{subcaption}
\usepackage{mwe}
\usepackage{threeparttable}
\usepackage{booktabs}
\usepackage{hyperref}
\usepackage[table]{xcolor}

\newcommand{\regrad}[0]{REGRAD\xspace}

\usepackage{amssymb}
\usepackage{bbding}
\usepackage{algorithm}
\usepackage{caption}
\usepackage[misc]{ifsym}

\title{\LARGE \bf
\regrad: A Large-Scale Relational Grasp Dataset for Safe and Object-Specific Robotic Grasping in Clutter
}

\author{Hanbo Zhang*, Deyu Yang*, Han Wang*, Binglei Zhao, Xuguang Lan \Letter, Jishiyu Ding, Nanning Zheng
\thanks{
$*$ Equally contributed.
}
\thanks{
Corresponding Author: Xuguang Lan.
Xuguang Lan is with the Institute of Artificial Intelligence and Robotics, the National Engineering Laboratory for Visual Information Processing and Applications, School of Electronic and Information Engineering,
        Xi'an Jiaotong University, No.28 Xianning Road, Xi'an, Shaanxi, China.
        {\tt\small xglan@mail.xjtu.edu.cn}
        }%
\thanks{
This work was supported in part by the key project of Trico-Robot plan of NSFC under grant No. 91748208, National Key Program of China No.2017YFB1302200,key project of Shaanxi province No.2018ZDCXL-GY-06-07, and NSFC No.61573268.}
}

\begin{document}

\maketitle
\thispagestyle{empty}
\pagestyle{empty}

\begin{abstract}

Despite the impressive progress achieved in robotic grasping, robots are not skilled in sophisticated tasks (e.g. search and grasp a specified target in clutter). Such tasks involve not only grasping but the comprehensive perception of the world (e.g. the object relationships). Recently, encouraging results demonstrate that it is possible to understand high-level concepts by learning. However, such algorithms are usually  data-intensive, and the lack of data severely limits their performance. In this paper, we present a new dataset named REGRAD for the learning of relationships among objects and grasps. We collect the annotations of object poses, segmentations, grasps, and relationships for the target-driven relational grasping tasks. Our dataset is collected in both forms of 2D images and 3D point clouds. Moreover, since all the data are generated automatically, { it is free to import new objects for data generation. We also released a real-world validation dataset to evaluate the sim-to-real performance of models trained on \regrad. Finally, we conducted a series of experiments, showing that the models trained on \regrad could generalize well to the realistic scenarios, in terms of both relationship and grasp detection.} Our dataset and code could be found at: \href{https://github.com/poisonwine/REGRAD}{https://github.com/poisonwine/REGRAD}.

\end{abstract}

\section{Introduction}
Robotic grasping is a fundamental problem in robotics. 
It plays a basic role in nearly all robotic manipulation tasks.
Particularly, visual perception is important for robotic grasping, since it can provide rich observations about the surroundings.
{Recently, with deep learning, robotic grasping has achieved impressive progress.
For example, the deep-learning methods \cite{redmon2015real, mahler2016dex, zhao2020regnet} achieved state-of-the-art performance on different benchmarks with surprisingly good generalization to unknown objects.}
However, grasping in realistic scenarios is usually target-driven,
{which often requires the understanding of high-level visual concepts, e.g., the object relationships.}
Though currently, to some extent, we can generate stable grasps in either scattered or cluttered scenes based on the advanced grasping algorithms, it is still hard for the robot to complete sophisticated manipulation tasks.

\begin{figure}[t] 
 \center{\includegraphics[width=0.42\textwidth]{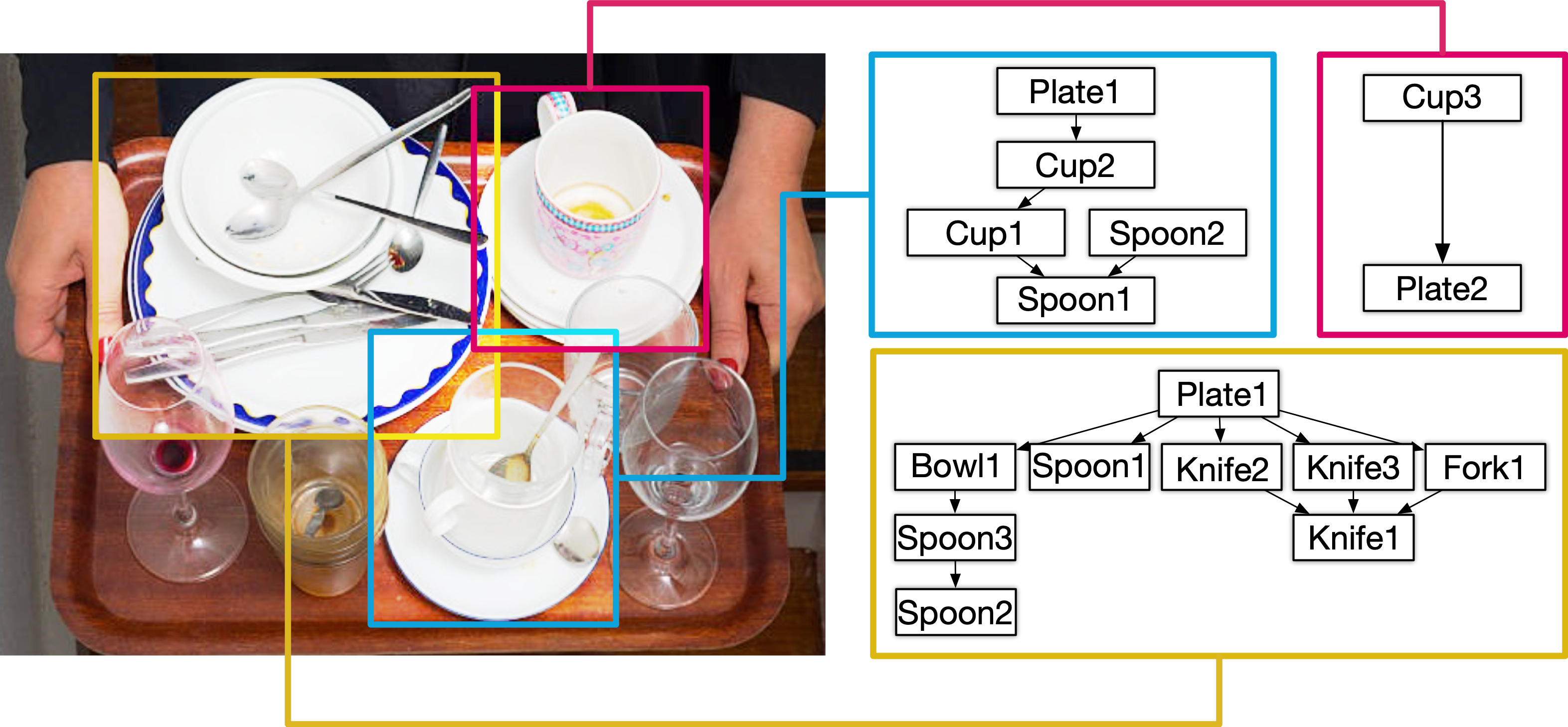}}   
 \caption{An example of a sophisticated target-driven grasping task in dense clutter.
 {\bf Left}: the working scenario. {\bf Right}: the correct grasping order among objects.
 Obviously, grasping without understanding the high-level visual concepts (e.g. relationships among objects and grasps) will result in a failure and even severe damages to other objects.
 } 
 \label{fig:motivation}
 \end{figure}

Suppose a common scene in our life as shown in Figure \ref{fig:motivation}.
If we want the robot to get the bottom plate for us, how can the robot {achieve it?}
Such a sophisticated task is obviously much harder than just finding a stable grasp of some object and raises two main challenges:
\begin{itemize}
\item {{\bf The understanding of object relationships} is necessary for inferring the correct grasping order shown on the right of Figure \ref{fig:motivation}.} Otherwise, it may cause irrevocable damages to other objects.
\item The robot is required to perform {\bf object-specific grasping} in dense clutter, which causes severe overlaps and occlusions among objects.
\end{itemize}

{ Previous works \cite{zhang2018visual, zhang2019roi} explored the manipulation relationship detection and object-specific grasping in clutter.
However, such deep learning methods are usually data-intensive \cite{lecun2015deep}.
The dataset, VMRD \cite{zhang2018visual}, containing only 200 object instances restricted their generality in real-world scenarios, and the human bias in VMRD exacerbates the overfitting since the model inevitably encoded the bias during learning.
}
To solve these problems, in this paper, we propose a novel, large-scale, and automatically-generated dataset.
By considering the relationships among objects and grasps, our dataset, called RElational GRAsp Dataset (\regrad), aims to build a benchmark for {relational} grasping in dense clutter.

 \begin{table*}[t]
\caption{Comparison with Related Datasets}
\label{comp}
\begin{center}
\begin{tabular}{c|c|c|c|c|c|c|c|c|c|c|c|c|c}
\hline
\bf \multirow{2}{*}{Dataset} & \bf Num    & \multirow{2}{*}{\bf Modality} & \bf Obj. & \bf Grasps & \bf Grasp &  \bf \multirow{2}{*}{Type}  & \bf Rel. & \bf 6D                & \bf Seg.  &  \bf Num & \bf Num & \bf Num & \bf Num \\ 
                                           & \bf Imgs &                                             & \bf /Img   & \bf /Img  & \bf Label   & \bf   &\bf Label      &\bf Pose                & \bf Label & \bf Cat. & \bf Obj. & \bf Rel. & \bf Grasps\\ 
\hline
Cornell                               &  1035         & RGB-D                                & 1                 &  $\sim$8    & Rect           &  Real            & -                  &   $\times$ &$\times$ &    -          &  240          & 0                 & 8K \\
Mahler et al.\cite{mahler2016dex}                      &   6.7M         & Depth                                 &  1                 &  1               & Rect           &  Sim            & -                  &  $\times$ & $\times$ & -          &  1500        & 0                 & 6.7M \\
Levine et al.\cite{levine2018learning}                      &   800K         & RGB-D                               & -                   &  1               & Rect           & Real & $\times$ & $\times$ & $\times$&-        &  -             & 0                 & 800K \\
Jacquard\cite{depierre2018jacquard}                           &   54K         & RGB-D                               & 1                  &  $\sim$20     & Rect           & Sim            & -                  &  $\times$     &  \checkmark&-         & 11K           & 0                 & 1.1M \\
VMRD\cite{zhang2018visual}                              &    4.7K        & RGB                                   &  $\sim$3     &  $\sim$20      & Rect        & Real & \checkmark & $\times$   & $\times$&31          &  $\sim$200          & 46K            & 100K  \\
\multirow{2}{*}{GraspNet\cite{fang2020graspnet}}                         &    \multirow{2}{*}{97K}         & \multirow{2}{*}{RGB-D}                               & \multirow{2}{*}{$\sim$10}       & \multirow{2}{*}{3$\sim$9M}  &Rect+  &  \multirow{2}{*}{Real} & \multirow{2}{*}{$\times$} & \multirow{2}{*}{\checkmark} & \multirow{2}{*}{\checkmark} &\multirow{2}{*}{-}              & \multirow{2}{*}{88}        & \multirow{2}{*}{0} & \multirow{2}{*}{1.2B} \\
& & & & & 6D & & & & & & & \\
\hline
\multirow{2}{*}{\bf \regrad}&  \multirow{2}{*}{\bf 900K} & \multirow{2}{*}{RGB-D}      &   \multirow{2}{*}{1$\sim$20} &          \multirow{2}{*}{1.02K} & Rect+        & \multirow{2}{*}{Sim} &  \multirow{2}{*}{\checkmark} &  \multirow{2}{*}{\checkmark} &  \multirow{2}{*}{\checkmark}&\multirow{2}{*}{55}            &      \multirow{2}{*}{50K}     &     \multirow{2}{*}{12M}   &  \multirow{2}{*}{100M}

\\
                                             &                   &                     &                   &                     &  6D   &                       &                  &                        &                   &                &                  &                  &\\
\hline
\end{tabular}
\end{center}
\end{table*}
 
 \begin{figure*}[t] 
 \vspace{-12pt}
 \center{\includegraphics[width=0.9\textwidth]{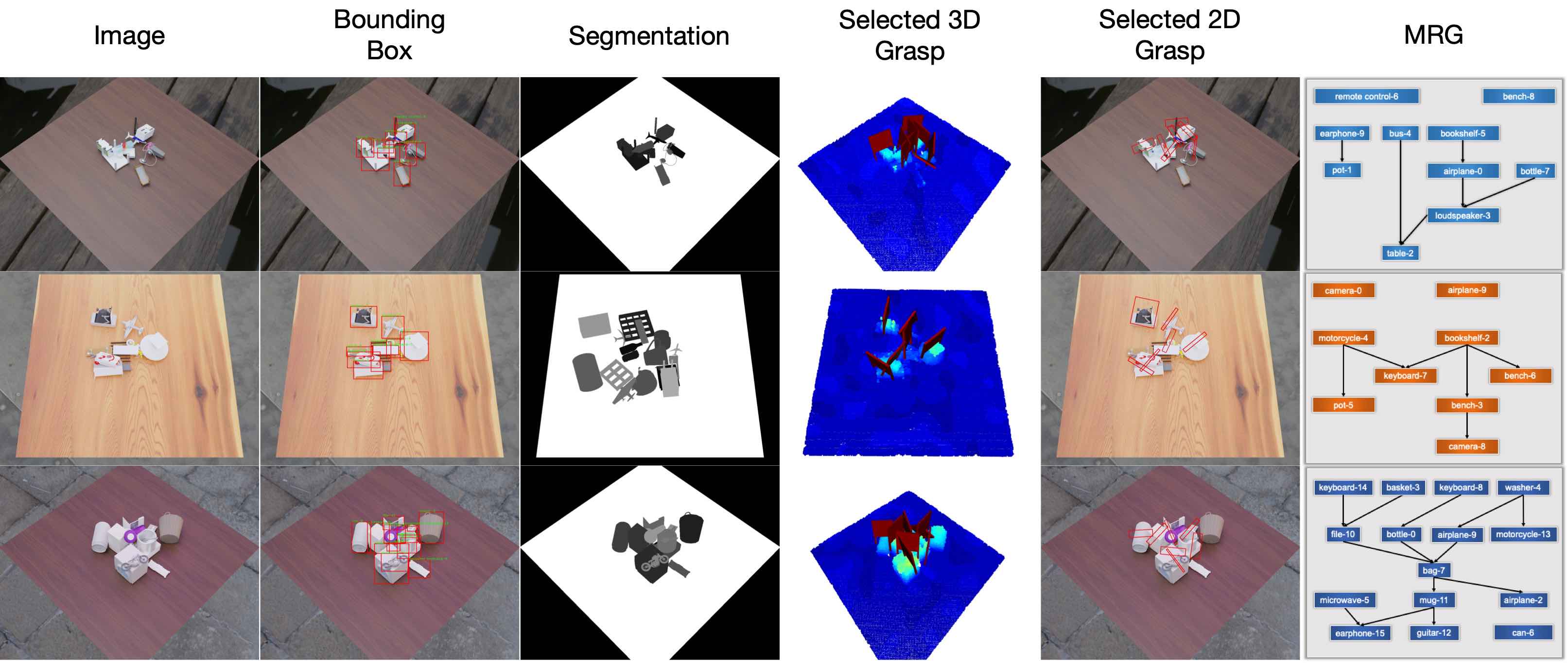}}   
 \caption{ Some examples of \regrad. The images are taken from 9 different views and the background is randomly generated. From left to right: {\bf Image}: the original RGB image of the scene; {\bf Bounding Box}: the minimal vertical rectangle to cover all pixels of the object; {\bf Segmentation}: the pixel-wise segmentation, with different pixel values representing different objects; {\bf Selected 3D Grasp}: the visualization of a selected subset of 6D grasp labels in the 3D working space; {\bf Selected 2D Grasp}: the rectangular (Rect.) grasp representation in the 2D image space, with center $(x, y)$, width $w$, height $h$, and the orientation $\theta$; {\bf MRG}: manipulation relationship graph representing the correct order of grasping.} 
 \label{fig:examples}
 \end{figure*}
Our dataset has the following features:

\begin{itemize}

	\item \textbf{More objects and categories}. Our dataset is built upon the well-known {ShapeNet dataset \cite{wu20153d, chang2015shapenet}, including 55 categories and }50K different object models.

	\item {{\bf Different kinds of modalities} including the depth images and point clouds, which are helpful for relationship detection, grasp synthesis, and sim-to-real transferring.}

	\item {{\bf Rich labels} including}:
		\begin{itemize}
			\item 6D pose of each object.
			\item Bounding boxes and segmentations on 2D images.
			\item Point cloud segmentations.
			\item Manipulation Relationship Graph (MRG) indicating the grasping order.
			\item Collision-free and stable 6D grasps of each object.
			\item Rectangular 2D grasps.
		\end{itemize}

	\item {{\bf Segregated training, validation, and test sets} including the unseen validation set and unseen test set, in which the objects belong to unknown categories.}

	\item {{\bf Multi-view data} which releases the assumption of single-view perception since in practice the robot could move around for more precise manipulation.}

	\item { {\bf Automatic data generation} in the physical simulator which will save much time to label the dataset and avoid the bias from the human labels. Currently, to add new objects to the datasets, we only need to scan the 3D models instead of re-collecting and labeling images. We also provide the open-source codes for dataset generation\footnote{\href{https://github.com/poisonwine/REGRAD}{https://github.com/poisonwine/REGRAD}}.}

\end{itemize}

Currently, our dataset contains 111.8k different scenes, including 46.9k training scenes, 32.1k validation scenes, and 32.8k test scenes.
{To validate the trained models with unknown objects, the validation set and the test set are split} into two parts: the \textit{seen} and \textit{unseen} parts.
Figure \ref{fig:examples} shows some examples of our dataset. 
{The quantitative comparison with other existing datasets could be found in Table \ref{comp}.}

{In summary, our contributions include:
\begin{itemize}
	\item We propose to automatically label the manipulation relationships by leveraging the physical simulators, avoiding the expensive manual labeling as well as the human bias, which exacerbates the overfitting.
	\item We contribute a large-scale relational grasping dataset named \regrad. As we know, it is the first large-scale relational grasping dataset.
	\item We validate the sim-to-real performance of both manipulation relationship detection and grasp detection on \regrad, and the results show that the models trained on \regrad could surprisingly transfer to realistic scenarios only by proper domain randomization. techniques.
\end{itemize}}

\begin{figure*}
        \centering
        
        \begin{subfigure}[b]{0.4\textwidth}   
            \centering 
            \includegraphics[height=4cm]{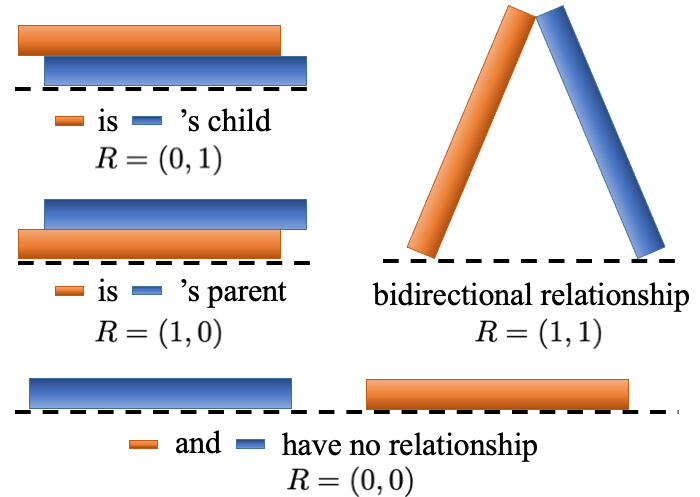}
            \caption{Manipulation relationship.}    
            \label{fig:reldef}
        \end{subfigure}
        \hfill
        \begin{subfigure}[b]{0.3\textwidth}
            \centering
            \includegraphics[width=\textwidth]{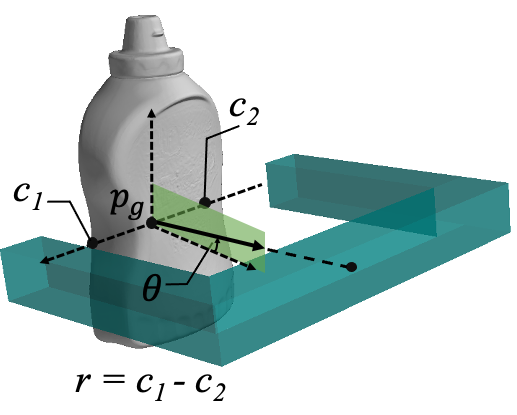}
            \caption{Grasp on point clouds.}    
            \label{fig:3dgraspdef}
        \end{subfigure}
        \hfill
        \begin{subfigure}[b]{0.25\textwidth}  
            \centering 
            \includegraphics[width=\textwidth]{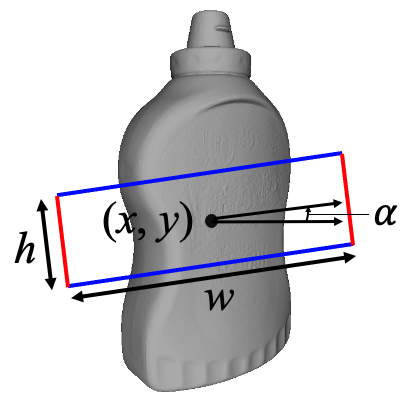}
            \caption{Grasp on images.}    
            \label{fig:2dgraspdef}
        \end{subfigure}
        
        \caption{Modeling of manipulation relationships and grasps in our dataset.}
\end{figure*}

\section {Related Work}

\textbf{Visual Relationship}~\cite{lu2016visual} is related to our manipulation relationship. It is defined as the relationship between objects inferred from visual images. 
Recently, a number of datasets have appeared, including the Visual Relationship Dataset~\cite{lu2016visual} and Visual Genome~\cite{krishna2017visual}.
Inspired by these works, Visual Manipulation Relationship Dataset (VMRD)~\cite{zhang2018visual} is proposed for robotic manipulation relationship detection.
Yet there are several drawbacks making models trained on VMRD not generalize well to realistic scenes: 1) VMRD only contains RGB images; 2) VMRD contains human biases when collected; { 3) The data is limited due to expensive manual labeling.}

\textbf{Spatial relation and support relation} are also related to the manipulation relationship. 
There are several works focusing on spatial relations in clutter~(e.g. \cite{rosman2011learning}). 
However, spatial relations cannot be directly used to guide the manipulation in clutter.
Another more relevant concept is named support relation~\cite{panda2013learning, mojtahedzadeh2015support, kartmann2018extraction}.
{ In support relation detection, it is usually assumed that objects could be properly represented by simple primitives, which restricts the application of such methods in the real world.
Moreover,} their performance will be affected by the upstream tasks (e.g. segmentation, geometry, and pose extraction, etc.).
By contrast, it is believed that with large-scale datasets, integrating deep learning into traditional robotic techniques may bring improvements to the generality of the existing methods.

\textbf{Robotic grasping} has been an actively investigated area in robotics for a long time\cite{bohg2013data}, but it still remains unsolved. 
Recently, with deep learning, researchers have achieved vast progress in robotic grasping~(e.g. \cite{lenz2015deep, redmon2015real, mahler2016dex, kumra2017robotic, levine2018learning, mahler2019learning, qin2020s4g}).
Such { methods heavily rely on} large-scale grasp datasets such as Cornell Grasp Dataset\footnote{http://pr.cs.cornell.edu/grasping/rect\_data/data.php}, 
DexNet\cite{mahler2016dex}, CMU Grasp Dataset\cite{pinto2016supersizing}, Jacquard Dataset\cite{depierre2018jacquard}, VMRD\cite{zhang2019roi}, and GraspNet\cite{fang2020graspnet}.
{
Particularly, object-specific robotic grasping in clutter is a practical but harder problem, which recently attracts researchers devoted to it~\cite{zhang2019roi, zhang2019multi, murali20206}.
For example, \cite{zhang2019multi} tries to detect grasps as well as the manipulation relationships with the model trained on VMRD \cite{zhang2018visual}.
However, it suffers from poor generality because of the limited dataset, which severely restricts its practicality.}

In this paper, we aim to build a new, large-scale, and automatically-generated grasp dataset considering all the above issues. 
We want to enable the robot to comprehensively percept the surroundings for grasp decision making.


\section{Modeling}

We include multi-modal labels of the objects both in point clouds and RGB images. For object detection and segmentation, we follow the traditional settings in previous works \cite{lin2014microsoft}. The representation of object grasps and relationships will be introduced in detail in the following sections.

%

\subsection{Visual Manipulation Relationships}
\label{sec:mrdef}
Following \cite{zhang2018visual}, \regrad focuses on the pair-wise manipulation relationships between objects.
{Specifically, the manipulation relationship between the object pair $(o_i, o_j)$ is a two-bit relationship $R=(R_{ij},R_{ji})$, with each bit denoting whether the relationship $R_{ij}$ or $R_{ji}$ exists or not.
If $R_{ij}=1$, the object $o_i$ should be grasped after $o_j$.
Therefore, there are totally 4 kinds of manipulation relationships in \regrad, which are demonstrated in Figure \ref{fig:reldef}.}


\subsection{Grasps on Point Clouds}
 
We demonstrate in Figure \ref{fig:3dgraspdef} how to define a grasp on the point cloud.
Typically, a grasp usually involves 4 key components: the grasp point, the orientation of the gripper, the approaching vector, and the grasp quality.
Therefore, a grasp on the point cloud can be formulated in Equation \ref{eq:3dgrasp}.
\begin{equation} \label{eq:3dgrasp}
	g_{pc} = (x,y,z,r_x,r_y,r_z,\theta,s)
\end{equation}
where $p_g=(x,y,z)$ defines the grasp point, $\textbf{r}=(r_x,r_y,r_z)$ defines the orientation of the gripper, $\theta$ represents the approaching direction given $p_g$ and $\textbf{r}$, and $s$ is {a score of quality} based on the antipodal measurement\cite{qin2020s4g}.

\subsection{Grasps on RGB Images}

Typically, the 2D grasps are parameterized by oriented rectangles following the same settings from \cite{jiang2011efficient}:
\begin{equation}
	g_{rect} = (x,y,w,h,\alpha)
\end{equation}
with $(x,y)$ representing the center of the grasp rectangle, $(w,h)$ denoting the width and height, and $\alpha$ being the counterclockwise orientation w.r.t. the horizontal axis of the image.
Concretely, the $(x,y)$ corresponds to the grasp point, and $(w,h)$ defines the gripper shape with $w$ being the distance between two fingers and $h$ being the width of each finger.
An example is given in Figure \ref{fig:2dgraspdef}.

\section{Dataset Generation}

\subsection{Preparation}

\subsubsection{Object Models}

\regrad is built on the basis of ShapeNet\cite{wu20153d}. 
We split the categories into \textit{seen} and \textit{unseen} parts.
The unseen categories will be used to generate data for validating and testing the performance of the trained models on completely unknown objects.

\subsubsection{Simulator}
In this paper, we use SAPIEN\cite{xiang2020sapien} as the simulator since it performs optimally compared with some other choices like PyBullet\cite{coumans2016pybullet} and Gazebo\cite{koenig2004design} considering the trade-off among the speed, authenticity, and robustness.

\subsection{Scene Generation}

All the scenes are generated automatically in the same procedure. 
In general, we first randomly sample a number of objects from all the object models. 
After that, we load them one by one into the simulator, resulting in a random clutter scene.

\subsubsection{Object Randomization} 
To load a model into the simulator, we need to define some parameters such as scale and friction. Specifically, to make the object graspable, we first sample a scale according to the longest side of the object's bounding box so that it is finalized in the range between 8cm and 20cm. For the friction, we sample the linear and angular damping coefficient between 1 and 1.5.
 
\subsubsection{Scene Randomization}
\begin{figure}[t] 
 \center{\includegraphics[width=0.45\textwidth]{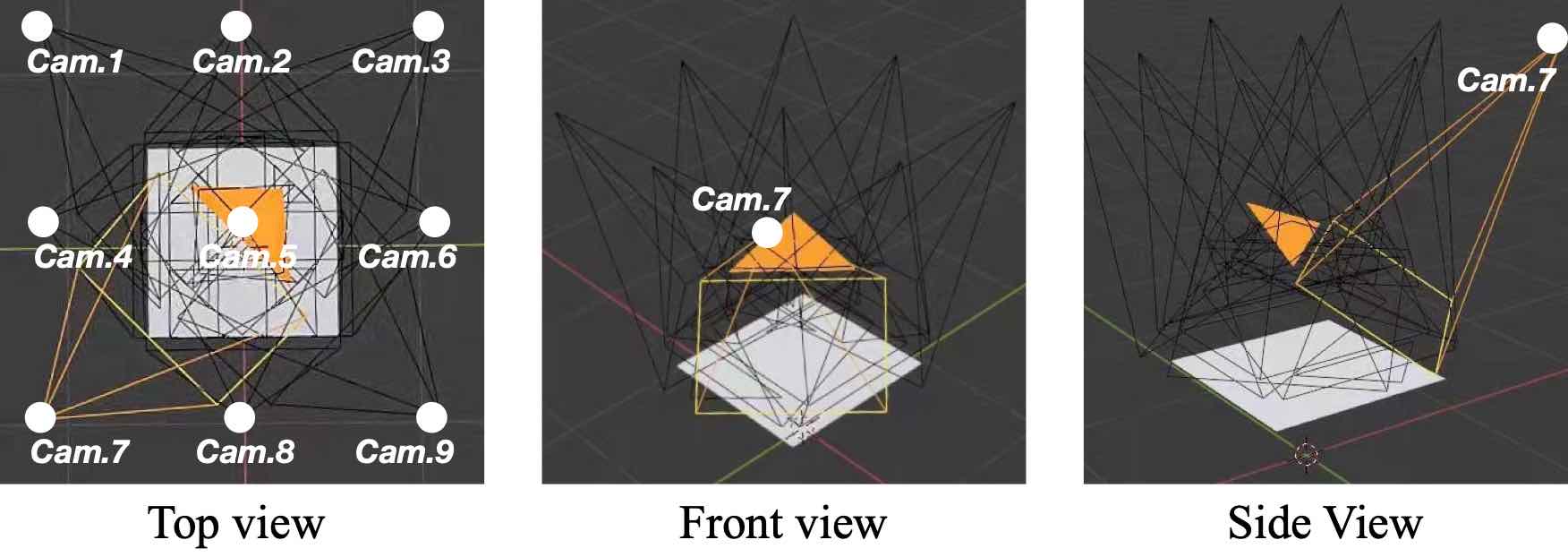}}   
 \caption{Three views of the 9 different camera poses.
 All 9 positions are at the same height above the table, and located at the 9 intersections of the $2\times 2$ square grids.} 
 \label{fig:campos}
 \end{figure}

To generate a scene, we need to define the scene background and some necessary parameters.
For the background, we use a random image to augment our dataset against the background noise for realistic tasks.
For the light, we randomly import 1-4 light sources around the table, enabling rich light conditions.
To load an object, we firstly sample an initial position above the table.
Then the object will drop down freely from the initial position.To avoid  severe collisions caused by high speed, we set the gravity acceleration of the simulator to 1/10  of the standard.
After loading all objects, we will record all scene parameters and object states so that we can completely recover it later.
For each scene, we will record the point clouds and RGB-D images from 9 different views as shown in Figure \ref{fig:campos}.
 
\begin{algorithm}[t] 
\caption{Algorithm of Automatically Generating MRG} 
\label{alg-mrg} 
\begin{algorithmic}[1] 
\Require 
The set of objects $O=\{o_i\}_{i=1}^{N_o}$; 
The object states $S=\{s_{o_i}\}_{i=1}^{N_o}$; 
A pose error threshold $\epsilon$;\\
Initialize the scene with the configuration recorded.
\For{$i=1,...,N_o$} 
\State Initialize the parent list of node $i$: $Par_i=\emptyset$
\For{$j=1,...,N_o$ and $j\neq i$}
\State Load $o_j$ using state $s_{o_j}$ in static mode.
\EndFor
\State Load $o_i$ using state $s_{o_i}$ in non-static mode.
\For{$j=1,...,N_o$ and $j\neq i$}
\State Delete $o_j$
\State Run several simulation steps.
\State Record $o_i$'s state $s_{o_i}'$
\If{$|s_{o_i}'-s_{o_i}|>\epsilon$}
$Par_i=Par_i\cup \{o_j\}$
\EndIf
Recover $o_i$ and $o_j$.
\EndFor
\EndFor 
\Return $MRG=\{Par_i\}_{i=1}^{N_o}$; 
\end{algorithmic} 
\end{algorithm}

\subsection{Manipulation Relationship Graph Generation}
\label{mrg-gen}

In this section, we will introduce the method to automatically label the manipulation relationship graph (MRG) for a given clutter scene.
The manipulation relationship indicates the correct grasping order {and the formal definition could be found in Section \ref{sec:mrdef}}.
Some examples of MRG are given in {the right column of} Figure \ref{fig:examples}.

The main idea to generate MRG is that we can separately find out all the parent nodes of each node based on physical simulation.
The procedure is summarized in Algorithm \ref{alg-mrg}.
Concretely, 
first of all, we initialize the scene using the recorded configuration without any objects (line 1).
To find out the parent nodes of $o_i$, we firstly load $o_i$ in non-static mode while all the other objects in static mode (line 4-6). 
The static mode means that the object cannot be affected by the simulation steps.
Therefore, moving the ancestor but not parent nodes of $o_i$ should not have a direct impact on $o_i$.
Only the moving of parents will affect the balance of $o_i$.
Thus, by iteratively moving all the other nodes, we can get a list of $o_i$'s parents (line 7-11).
Finally, the parent lists of all nodes will result in a complete MRG for the given scene.

\subsection{Grasp Generation}

Basically, we follow the method introduced in \cite{qin2020s4g} to generate 3D grasps and then follow \cite{yang2019task} to project them onto RGB images to get 2D rectangular grasps.

\label{grasp-gen}

\subsubsection{3D Grasps}
 

For grasp generation, we follow the flow of sample-first-then-filter.
To be specific, we firstly sample a set of grasp candidates around each object {by the following steps:
\begin{itemize}
	\item {\bf Point sampling} which samples a subset of points belonging to the object;
	\item  {\bf Depth sampling} which samples grasps with different depths along the direction of the surface normal at each point;
	\item {\bf Rotation angle sampling} which samples different rotation angles around the surface normal at each point;
	\item {\bf Inclination angle sampling} which samples different inclination angles around the vector between two fingers, namely different $\theta$s as shown in Figure \ref{fig:3dgraspdef}.
\end{itemize}
The sampled grasp pose of the gripper determines the approaching vector, which is within the gripper plane and perpendicular to the vector between two fingers, as well as the contact points when executing the grasping actions.
After that, we conduct the static collision checking, which means that we only consider the final pose of the gripper without movement, and filter out all grasps with collisions.
Finally, we assign an analytically-computed antipodal score and a center score to measure the quality of each grasp.}


{To get the \emph{antipodal score} $c_{a}$, we follow S$^4$G~\cite{qin2020s4g}.}
Besides, we assume that the high-quality grasps are usually located near the center of mass.
Given the assumption of uniform mass distribution, the mass center usually coincides with the geometric center.
Therefore, we assign the \emph{center score} $c_{c}$ describing the distance from the grasp to the geometric center of the object:
$$
c_{c} =  \frac{d_{max}-d}{d_{max}-d_{min}}
$$
where $d$ is the distance from the grasp to the object center, $d_{max}$ and $d_{min}$ are the maximum and minimum distance between any grasp and its corresponding object center. 

\subsubsection{2D Grasps}

Our 2D rectangular grasps on RGB images are generated automatically using the projection of the 3D grasps from the point clouds.

To generate the 2D grasps on RGB images, first and foremost, we assume that the extrinsic and intrinsic parameters of the camera are available, which always holds in the simulator. 
Given a grasp representation from point clouds, we can easily obtain the grasp pose of the gripper.
Provided the gripper pose, we define the two surfaces of the two fingers facing inward as the contact surface.
Then we record the two edges at the bottom of the two contact surfaces and project them to the image frame through the camera parameters, resulting in a grasp parallelogram.
The final grasp rectangle is defined as the minimum bounding rectangle of the grasp parallelogram.
Finally, to discard inferior grasps, we filter out the grasp rectangles projected from the 3D grasp poses  that have an angle larger than $30^{\circ}$ with the camera.

\section{Experiments}

To validate the effectiveness of \regrad, we run experiments on manipulation relationship detection and grasp detection on it.
{Results show that models trained on \regrad could generalize well to realistic scenarios.}

\subsection{Manipulation Relationship Detection}

\begin{table*}[t]
\caption{Manipulation Relationship Detection}
\vspace{-6pt}
\label{table:reldet}
\begin{center}
\begin{tabularx}{1\textwidth}{p{1.5cm}p{2.8cm}ccccccccccccc}

\toprule

& & \multicolumn{12}{c}{\bf Performance}\\
\cmidrule(r){3-6}  \cmidrule(r){7-10} \cmidrule(r){11-14}

\bf Metrics & \bf Training Data &  \multicolumn{4}{c}{\bf REGRAD-seen-val-500} &\multicolumn{4}{c}{\bf REGRAD-unseen-val-500}&\multicolumn{4}{c}{\bf Sim-to-real Transfer}\\

\cmidrule(r){3-3} \cmidrule(r){4-4} \cmidrule(r){5-5} \cmidrule(r){6-6} \cmidrule(r){7-7} \cmidrule(r){8-8} \cmidrule(r){9-9} \cmidrule(r){10-10} \cmidrule(r){11-11} \cmidrule(r){12-12} \cmidrule(r){13-13} \cmidrule(r){14-14}
 & & P & C & N & Mean & P & C & N & Mean & P & C & N & Mean \\
\midrule
\bf Recall & \bf VMRD-4683 & 17.59 & 22.55 & 95.78 & 45.31 & 16.48 & 21.35 & 95.46 & 44.43 & 22.82 & 26.97 & 97.02 & 48.94\\
&\bf REGRAD5K-RGB &\bf 60.88 & \bf 60.87 & 95.82 & \bf 72.52 & \bf 55.24 & \bf 54.32 & 95.00 & \bf 68.19 & \bf 90.04 & \bf 73.03 & 92.34 & \bf 85.14 \\
& \bf REGRAD5K-Depth & 59.11 & 55.90 &\bf 97.62 & 70.88 & 52.27 & 48.58 &\bf 96.93 & 65.93 & 77.18 & 61.83 & \bf 97.99 & 79.00\\

\midrule
\bf Precision & \bf VMRD-4683 & 21.27 & 19.86 & 95.67 & 45.60 & 21.44 & 19.57 & 95.00 & 45.34 & 25.94 & 22.49 & 97.17 & 48.53 \\
&\bf REGRAD5K-RGB & 44.22 & 43.30 & \bf 98.12 & 61.88 & 39.82 & 39.46 & \bf 97.58 & 58.95 & 29.05 & 27.16 & \bf 99.74 & 51.98 \\
& \bf REGRAD5K-Depth & \bf 54.76 & \bf  57.56 & 97.77 & \bf 70.03 & \bf 48.71 & \bf 50.11 & 97.07 &\bf  65.30 & \bf 50.00 & \bf 57.53 & 99.19 & \bf 68.91\\

\midrule
\bf Image Acc & \bf VMRD-4683 & \multicolumn{4}{c}{13.4}& \multicolumn{4}{c}{15.5}& \multicolumn{4}{c}{16.1} \\
&\bf REGRAD5K-RGB & \multicolumn{4}{c}{22.2}& \multicolumn{4}{c}{19.6}& \multicolumn{4}{c}{17.6}\\
& \bf REGRAD5K-Depth & \multicolumn{4}{c}{\bf 27.4}& \multicolumn{4}{c}{\bf 24.5}& \multicolumn{4}{c}{\bf 40.7}\\

\bottomrule
\end{tabularx}
\vspace{-13pt}
\end{center}
\end{table*}

\begin{table}[t]
\caption{Comparison with VMRD}
\vspace{-6pt}
\label{table:relcomp}
\begin{center}
\begin{tabularx}{1\columnwidth}{l@{\hspace{30pt}}c@{\hspace{30pt}}c@{\hspace{30pt}}c}
\toprule
 & VMRD~\cite{zhang2018visual} & \bf REGRAD & Times \\
\midrule
None & 24.9K & \bf 11.38M & \bf $\sim$455 \\
Parent &13.3K & \bf 430.7K &  \bf $\sim$33\\
Child &13.3K & \bf 430.7K &  \bf $\sim$33\\
Bidirectional &  - &\bf 81.6K & -\\
\midrule
Total & 51.5K &\bf 12.32M &  \bf $\sim$240\\
\bottomrule
\end{tabularx}
\vspace{-8pt}
\end{center}
\end{table}

\begin{table}[t]
\centering
\caption{Comparison with REGNet Dataset}
\begin{threeparttable}
\begin{tabularx}{\columnwidth}{m{0.2\columnwidth}>{\centering}m{0.2\columnwidth}>{\centering}m{0.2\columnwidth}c}
\toprule[1pt]
& \bf MAS & \bf MNG & \bf MNG(0.5) \\
\midrule
REGNet\cite{zhao2020regnet} &  0.5623 & 294 & 196  \\ 
REGRAG & 0.5000  &  1040 & 503\\ 
\bottomrule[1pt]
\end{tabularx}
\end{threeparttable}
\label{table:count6d}
\end{table}

{
We firstly compare the performance of manipulation relationship detection on \regrad and VMRD.
The difference between these two datasets in terms of manipulation relationships is shown in Table \ref{table:relcomp}.
To conduct the experiments, we test the performance on three validation datasets with comparison to VMRD~\cite{zhang2018visual}:
\begin{itemize}
	\item {\bf \regrad-seen-val-500}: the first 500 scenes of \regrad unseen-val set, consisting of 4500 images including different object instances of known categories.
	\item {\bf \regrad-unseen-val-500}: the first 500 scenes of \regrad unseen-val set, consisting of 4500 images of unknown objects.
	\item {\bf Sim-to-real transfer}: 200 realistic scenarios collected and labeled with real objects and a Kinect DK camera.
\end{itemize}
To do so, we implement the Visual Manipulation Relationship Network \cite{zhang2018visual} on \regrad, and train the model using the same settings.
To make our model compatible with VMRD, we ignore the bidirectional relationship in \regrad.}

{
Note that to facilitate the sim-to-real transfer, we apply domain randomization techniques, i.e., a series of data augmentations are applied to the input images before we send them into the neural network.
We conduct experiments using both RGB and depth images as the inputs.
For RGB images, the standard augmentation, which is the same as the one in \cite{zhang2018visual}, is performed before the image is input to the neural network.
While for depth images, we impose the augmentation to make the augmented images similar to the real ones.
In detail, our augmentation includes:
\begin{itemize}
	\item Random Gaussian noise with standard derivation from 0 to 2.5.
	\item Random granularity of 2.5 pixels to 5 pixels.
	\item Random Gaussian blurring with different kernel sizes.
	\item Random black holes on the edge of each object implemented using image high-pass filter.
	\item Random strip noise by adding a gray strip image onto the original image.
	\item Random brightness by multiplying the depth image with a random factor ranging from 0.6 to 1.2.
\end{itemize}
The first three augmentations will be executed with a random number of iterations for each depth image.
}

{
The evaluation metrics include recall, precision, and image accuracy as in \cite{zhang2018visual}.
However, since we are willing to compare the relationship performance on unknown objects, here we ignore the object detection part and use the ground truth bounding box for pure relationship classification.
We respectively evaluate the performance of different kinds of relationships including parent relation ({\bf P}), child relation ({\bf C}), and no relation ({\bf N}), 
and the results are shown in Table \ref{table:reldet}. 
We can conclude that:
\begin{itemize}
	\item The model trained on RGB images has a higher recall. This is because the RGB-based model is more aggressive, tending to misclassify the object pair with no relation to parent or child relation.
	\item The model trained on depth images has a much higher precision and image accuracy on all three validation sets. Also, it transfers better to the real scenarios, with nearly no performance loss and much higher precision, while the model trained on RGB images suffers from a severe sim-to-real performance loss.
	\item The prediction of no relation is more reliable than the one of parent and child relation due to data imbalance.
	\item The model trained on VMRD cannot generalize well to new object instances, showing the worst generalization performance.
\end{itemize}
}

\subsection{Grasp Detection}

We test the 3D grasp detection performance on \regrad based on the state-of-the-art grasp detector REGNet~\cite{zhao2020regnet} following the same settings.
To evaluate whether the model trained on \regrad can generalize well, we also conduct the cross-dataset validation by deploying two models on \regrad and REGNet datasets \cite{zhao2020regnet}. 
Following \cite{zhao2020regnet}, we use two metrics to evaluate the grasp performance: collision-free ratio and antipodal score, which describe the possibility of each grasp having no collision with objects and the force-closure property respectively. 

Table \ref{table:count6d} shows the differences between \regrad and REGNet datasets. \textit{MAS} denotes the mean antipodal score, \textit{MNG} represents the mean number of grasps, and \textit{MNG(0.5)} is MNG with an antipodal score larger than 0.5. The results are illustrated in Table \ref{table:6dres}, where \textit{Cf-R} is the average collision-free rate and \textit{AS} is the average antipodal score provided the detected grasps. We can conclude that:
\begin{itemize}
    \item \regrad has denser labels in each scene than REGNet dataset.
    \item The model trained on \regrad generalizes well to REGNet dataset, with low collision rate and high antipodal score.
    \item The model trained on REGNet only works well on REGNet dataset and suffers from a sharp drop of performance when transferred to \regrad.
    \item \regrad is harder than REGNet dataset. Both models perform worse on \regrad compared to REGNet dataset, since 1) the large number of non-convex models in \regrad leads to a lower antipodal score, and 2) the scenarios in \regrad are more complicated.
\end{itemize}

{
We also conducted the real-robot experiments to execute the detected grasps in real-world scenarios, and the results are shown in Table \ref{table:realrobotgrasp}. 
In detail, we run the models in 10 different scenarios, including 6-12 objects which are stacked together.
The grasping procedure will end until there is no available grasp detected anymore.
We can see that benefitting from the large-scale dataset, the model trained on \regrad has a higher grasp success rate (86.63\%) compared to the model trained on REGNet dataset (79.34\%) in the real world.
}

\begin{table}[t]
\centering
\caption{Cross-dataset Grasp Detection Performance}
\begin{threeparttable}
\begin{tabularx}{\columnwidth}{>{\centering}m{0.22\columnwidth}>{\centering}m{0.22\columnwidth}>{\centering}m{0.22\columnwidth}c}
\toprule[1pt]
\bf Train & \bf Test & \bf Cf-R & \bf AS\\
\midrule
REGNet\cite{zhao2020regnet}  & REGNet\cite{zhao2020regnet} & 82.11\% & 0.5690 \\
REGNet\cite{zhao2020regnet}  & REGRAD & 78.53\% & 0.3741 \\ 
REGRAD & REGNet\cite{zhao2020regnet}  & 81.32\% & 0.5651 \\
REGRAD & REGRAD & 79.32\% & 0.4353 \\ 
\bottomrule[1pt]
\end{tabularx}
\end{threeparttable}
\label{table:6dres}
\end{table} 

\begin{table}[t]
\centering
\caption{Real-robot Grasping Performance}
\begin{threeparttable}
\begin{tabularx}{\columnwidth}{>{\centering}m{0.3\columnwidth}>{\centering}m{0.3\columnwidth}c}
\toprule[1pt]
\bf Training Data & \bf Success Rate & \bf Complete Rate\\
\midrule
 REGNet\cite{zhao2020regnet}  & 79.34\% & 96.00\% \\
 REGRAD  & 86.63\% & 96.17\% \\ 
\bottomrule[1pt]
\end{tabularx}
\end{threeparttable}
\label{table:realrobotgrasp}
\end{table}

\section{Conclusions and Future Work}

In this paper, we contribute a large-scale and automatically-generated {relational grasping} dataset, namely \regrad (RElational GRAsp Dataset). 
Our dataset is the first one targeting comprehensive perceptual tasks in robotic manipulation, including object and pose detection, segmentation, target-driven grasping, and relationship understanding.
Along with the dataset, we also provide a principled way to automatically generate {relation labels} for deep-learning-based manipulation, meeting the {data-intensive} models.
We also implemented a series of state-of-the-art algorithms on \regrad, serving as the baselines.

For future work, we aim to expand \regrad using object captions for human-robot interaction. 
The model-wise captions from ShapeGlot~\cite{ achlioptas2019shapeglot} can be used for generating image-wise dense captions automatically by leveraging the spatial relationships and the relative positions among objects.
We also intend to expand \regrad with {
more sophisticated relationships.
Notably, the current definition of manipulation relationship is simple, object-agnostic, yet sufficient for most cases.
However, in our practice, the manipulation relationships among objects are subtle.
The relationship should sometimes depend on the attributes of the objects, e.g., the mass distribution and fragility.
For example, when a pen (instead of a cup) is on top of a book, we tend to grasp the book directly since this grasping action is harmless. 
Therefore, taking into consideration of object attributes in the MRG is promising to improve the efficiency of the current algorithm.
Also, the relationships could depend on the action itself \cite{sun2014object, mo2021o2o}.
In our current formulation, the manipulation relationship is only considered in the context of grasping.
However, by formulating the tuple of object-action-object, the MRG could support more flexible manipulations.

Finally, there will be a wide range of possible directions based on \regrad, for example:

\subsubsection{Object-Agnostic Manipulation Relationship Detection}
As mentioned, despite its simplicity and suboptimality, the current object-agnostic definition of manipulation relationship is sufficient for most cases.
Yet, its robust detection still remains unsolved, especially for unknown objects.
Some visual relationship detection algorithms raise up interesting perspectives for object-agnostic relationship detection~\cite{zhang2017visual, yang2018shuffle, hung2020contextual}.
Such ideas (e.g. introducing network structure bias to alleviate the object-specific features and adversarial training) are hopeful to be further explored to detect manipulation relations between unknown objects.
Besides, support relation analysis could generalize well to unknown objects~\cite{panda2013learning, mojtahedzadeh2015support, kartmann2018extraction, grotz2019active, paus2021probabilistic} provided the physical properties like mass distribution and friction coefficient~\cite{mojtahedzadeh2015support} or assumptions that objects could be properly approximated by simple geometric primitives~\cite{panda2013learning,kartmann2018extraction, grotz2019active, paus2021probabilistic}.
To follow such ideas in the future, the main challenge is how to get rid of the strong reliance on physical properties and geometric assumptions.

\subsubsection{Target-driven Grasp Detection in Clutter}
Target-driven grasping should be based on robust object-specific grasp detection.
To achieve so, one possible way is utilizing object-specific features to detect grasps in point clouds.
Fortunately, recent advances in deep learning show promising ways for extracting point cloud features (e.g.~\cite{qi2017pointnet, ni2020pointnet++, guo2020pct, wang2021spatial}).
Based on the advanced point cloud feature extractor, we will try to solve the following problems of target-driven grasping in dense clutter in the future:
\begin{itemize}
    \item Develop object-specific grasping feature extractors.
    \item Segment unknown objects accurately in clutter.
    \item Design robust object-specific grasp detector.
    \item Filter out grasps that are not for the targets.
    \item Avoid potential collisions when grasping.
\end{itemize}

\subsubsection{Sim-to-Real Transfer Learning}
Since \regrad is generated automatically using the physical simulator, there exists a reality gap between the training data and realistic scenarios.
Therefore, facing the challenges of sim-to-real is a practical and important issue.
Currently, there are mainly three ways toward solving this problem: \textit{fine-tune}, \textit{domain randomization}~\cite{tobin2017domain}, and \textit{domain adaptation}~\cite{wang2018deep}. 
Fine-tune means training on simulative data while followed by a fine-tune stage using real-world data, which would dramatically reduce the data amount required by training from scratch (e.g.~\cite{rusu2017sim, chebotar2019closing}). 
However, though the requirements are relaxed, it still needs a real-world dataset with a considerable size, which would be time-consuming to collect.
Domain randomization usually defines a diversified data space in the simulator, in which the real-world data are assumed to be.
By contrast, the key idea of domain adaptation is to build a shared space when given one or more \textit{source} domains (e.g. the simulative data) and a \textit{target} domain (the realistic data).
In this paper, we have already verified the vanilla domain randomization techniques, which successfully help our models, especially the depth-based model, transfer to real-world scenarios.
In the future, we could also consider the domain adaptation method based on adversarial learning~\cite{wang2020progressive, xu2020adversarial} and contrastive learning~\cite{liu2021domain, thota2021contrastive} to improve the sim-to-real performance.
}

We believe that \regrad will provide robotics researchers with more chances to face the challenges of complex robotic manipulation tasks.

\addtolength{\textheight}{0cm}   




\bibliographystyle{IEEEtran}
\bibliography{ICRA2021}

\end{document}